# Approximate Decomposition: A Method for Bounding and Estimating Probabilistic and Deterministic Queries

David Larkin

## Abstract

In this paper, we introduce a method for approximating the solution to inference and optimization tasks in uncertain and deterministic reasoning. Such tasks are in general intractable for exact algorithms because of the large number of dependency relationships in their structure. Our method effectively maps such a dense problem to a sparser one which is in some sense "closest". Exact methods can be run on the sparser problem to derive bounds on the original answer, which can be quite sharp. On one large CPCS network, for example, we were able to calculate upper and lower bounds on the conditional probability of a variable, given evidence, that were almost identical in the average case.

## 1 INTRODUCTION

Belief networks [Pearl, 1988] are a widely used formalism for reasoning with uncertainty in Artificial Intelligence. Unfortunately, basic computations on belief networks, such as calculating the probability of a query variable given evidence, or finding the probability of the most probable explanation consistent with a certain variable value given evidence, are NP-hard. Clique tree propagation [Jensen *et al.*, 1990] is the most popular exact algorithm, which requires time and space exponential in the treewidth of the network's interaction graph. Variable elimination [Zhang and Poole, 1994; Dechter, 1999] is a simplified formulation of this method, which only computes the answer to one query, but which is easier to derive and understand. Its complexity is also exponential in the treewidth. Approximation algorithms include iterative belief propagation [Pearl, 1988], stochastic simulation [Pearl, 1988], variational methods [Jordan *et al.*, 1999], and mini buckets [Dechter and Rish, 1997]. Iterative belief propagation provides an estimate of the exact answer, if it converges, but there is no guarantee of accuracy. Stochastic simulation also provides an estimate, along with a level of confidence, but it can be expensive and there is always some chance that the answer is significantly inaccurate. Variational methods provide guaranteed bounds, but must be tailored to specific classes of networks. They cannot be applied systematically to general networks. Mini buckets is a simple algorithm that works on general networks, which also provides guaranteed bounds, but in practice these are too loose to be useful. Mini buckets can also provide bounds for the MAX-CSP problem, where it compares well with state of the art methods [Kask and Dechter, 2001].

In this paper, we introduce an algorithm called approximate decomposition which is designed to provide tight guaranteed bounds on probabilistic and deterministic queries. It works by bounding a large complex function with a combination of simpler ones, such that the expected loss of accuracy in a query involving the function will be minimized.

The paper is divided into several parts. Following this introduction, we define basic concepts in section 2. In section 3 we describe the approximate decomposition algorithm. Finally in section 4 we describe our empirical results, and in section 5 we conclude and discuss possibilities for future research.

## 2 BASIC CONCEPTS

In this section we review basic concepts which will be important in succeeding sections.

### 2.1 BELIEF NETWORKS

A belief network is a tuple $(X, D, G, P)$ where $X$ is a set of $n$ variables $\{X_1, X_2, ... X_n\}$ and $D$ is a set of variable domains $\{D_1, ... D_n\}$. We use $x_i$ to denote a value from $X_i$'s domain $D_i$, $x$ to represent a vector $(x_1, x_2, ...x_n)$ of values for all variables, and $x_S$ to represent a choice of values for a subset $S$ of $X$. $G$ is a directed acyclic graph, and $P$ is a set of conditional probability tables $\{P(X_i|pa_i)\}$, also called CPTs. The parent set $pa_i$ of $X_i$ is the set of nodes which are sources of arcs pointing into $X_i$. $P(X_i = x_i|pa_i = x_{pa_i})$ is the conditional probability of $X$ taking the value $x_i$, given that its parents are assigned $x_{pa_i}$. The semantics of the network is a factorized joint probability distribution over $X$: $P(X = x) = \prod_{i=1}^n P(X_i = x_i|pa_i = x_{pa_i})$, where $x_i$ and $x_{pa_i}$ are consistent with $x$.

## 2.2 BELIEF NETWORK TASKS

The belief inference task is to compute the conditional probability $P(X_q = x_q|E = x_E)$ of a query variable $X_q$ for any of its values $x_q$, given some evidence $E = x_E$. Bayes's rule allows us to expand this as $P(X_q = x_q \wedge E = x_E)/P(E = x_E) = \alpha P(X_q = x_q \wedge E = x_E)$, where $\alpha$ is a normalizing constant. By the definition of the network $P(X_q = x_q \wedge E = x_E) = \sum_{\{X-(E\cup X_q)\}} \prod_{i=1}^n P(X_i = x_i|pa_i = x_{pa_i})|_{E=x_E}$, where the sum is over all assignments to the subscripted set of variables and $f(x)|_{E=x_E}$ is $f(x)$ when $x$ is consistent with $E = x_E$ and zero otherwise.

The MPE task is to find the probability of the most probable assignment $X = x$ which is consistent with the evidence $E = x_E$ and any value of a query variable $X_q$. Formally, again by the network definition, this is $\max_{\{X-(E\cup X_q)\}} \prod_{i=1}^n P(X_i = x_i|pa_i = x_{pa_i})|_{E=x_E}$.

## 2.3 MAX-CSP

In the MAX-CSP problem, we are given a set of constraints $C = \{C_1, ..., C_m\}$ over the variables $X$. Each constraint $C_i$ is a function defined on a subset of $X$ called its scope. It maps assignments to the scope that satisfy it to 0, and unsatisfying assignments to 1. The best solution violates the minimum number of constraints. Given a query variable $X_q$, the goal is to find the cost of the best solution consistent with any of its values, or $\min_{\{X-X_q\}} \sum_i C_i$.

## 2.4 VARIABLE ELIMINATION

Variable elimination [Zhang and Poole, 1994; Dechter, 1999] is an exact algorithm for probabilistic and deterministic reasoning. It can be applied to any of the tasks mentioned above. The basic operation is to transform an expression $\otimes_{\{X_1,...,X_k\}} \odot_{i=1}^m f_m$ to an expression $\otimes_{\{X_1,...,X_{k-1}\}} \odot_{i=1}^{m'} f_i'$ which is equivalent and does not mention $X_k$. This transformation is called *eliminating* $X_k$. We assume that $\otimes$ and $\odot$ are commutative and associative binary operations over the real numbers, and that $\otimes_{X_i}(f_j \odot f_k) = f_j \odot (\otimes_{X_i} f_k)$ if $f_j$ does not depend on $X_i$ (in other words, $\odot$ distributes over $\otimes$). The transformation is done by writing $\odot_{i=1}^m f_m$ as $(\odot_{i=1}^h f_i) \odot (\odot_{j=h+1}^m f_j)$ where only functions $h+1$ to $m$ depend on $X_k$ (renumbering if necessary). We can then write $\otimes_{\{X_1,...X_{k-1}\}} \otimes_{X_k} (\odot_{i=1}^h f_i) \odot (\odot_{j=h+1}^m f_j)$ as $\otimes_{\{X_1,...X_{k-1}\}} (\odot_{i=1}^h f_i) \odot (\otimes_{X_k} \odot_{j=h+1}^m f_j)$. We then define a new function $\lambda = \otimes_{X_k} \odot_{j=h+1}^m f_j$ which does not depend on $X_k$, and the expression $\otimes_{\{X_1,...X_{k-1}\}} \odot_{i=1}^h f_i \odot \lambda$ is then the desired result. After $X_k$ to $X_1$ have been eliminated, we will be left with a constant or a function on the uneliminated variables which is equivalent to the original expression, but which can be evaluated in $O(1)$ time.

For belief inference, we let $\otimes$ be summation, and $\odot$ becomes product. For the MPE task, $\otimes$ is the max function, and $\odot$ is product. Initially the CPTs in the network are simplified by instantiating the evidence variables with their values. Then $f_1, ..., f_m$ will be the CPTs, and all variables will be eliminated but the query. For MAX-CSP, $f_1, ..., f_m$ are the constraints, $\otimes$ is min, $\odot$ is summation, and again all variables but the query will be eliminated. The result of variable elimination then will be a unary function on the query variable, giving the desired answer for each of its values.

## 3 APPROXIMATE DECOMPOSITION

In this section we describe the approximate decomposition algorithm.

### 3.1 THE MAIN ALGORITHM

A collection of functions $\{f_1, ..., f_m\}$ over a set of variables $X$ can be represented by an undirected graph. There is a node for each variable in $X$, and a pair of variables are connected by an edge if they both appear in the scope of $f_i$, for some $i$. The undirected graph representing the CPTs of a belief network, called the moral graph, is found by connecting all parents of a common child in the network's DAG and undirecting the edges. See figure 1 for an example.

When a variable $X_k$ is eliminated by variable elimination, all functions mentioning it are removed and a new function is defined on all of its neighbors. This corresponds to deleting $X_k$ from the graph and connecting all neighbors. The cost of calculating the new function is exponential in the number of neighbors. In general, as variables are eliminated, the graph of the remaining problem gets denser and denser, until all variables have more than $i$ neighbors, for some fixed affordable complexity limit $i$. At that point the algorithm cannot proceed without an unacceptable cost.

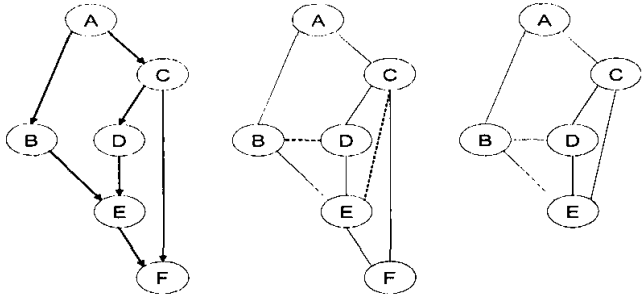

Figure 1: Example of Approximate Decomposition. *Left:* Input network. *Middle:* Moralized network graph. *Right:* After eliminating $F$.

| $P(A)$ : | $P(B|A)$ : | $P(C|A)$ : |
|---|---|---|
| $P(\overline{A}) = .6$ | $P(\overline{B}|A) = .3$ | $P(\overline{C}|A) = .4$ |
| $P(A) = .4$ | $P(B|\overline{A}) = .7$ | $P(C|\overline{A}) = .6$ |
| | $P(\overline{B}|A) = .5$ | $P(\overline{C}|A) = .8$ |
| | $P(B|A) = .5$ | $P(C|A) = .2$ |

| $\lambda(B,C)$ : | $\lambda_1(B), \lambda_2(C)$ : | $\lambda_1(B) \cdot \lambda_2(C)$ : |
|---|---|---|
| $\lambda(\overline{B},\overline{C}) = .23$ | $\lambda_1(\overline{B}) = 1$ | $\lambda_1(\overline{B}) \cdot \lambda_2(\overline{C}) = .23$ |
| $\lambda(\overline{B},C) = .15$ | $\lambda_1(B) = 1.41$ | $\lambda_1(\overline{B}) \cdot \lambda_2(C) = .207$ |
| $\lambda(B,\overline{C}) = .33$ | $\lambda_2(\overline{C}) = .232$ | $\lambda_1(B) \cdot \lambda_2(\overline{C}) = .33$ |
| $\lambda(B,C) = .29$ | $\lambda_2(C) = .207$ | $\lambda_1(B) \cdot \lambda_2(C) = .29$ |

Figure 2: Approximating $\lambda(B,C)$ with $\lambda_1(B) \cdot \lambda_2(C)$

The *width* of a graph is a measure of its density. It is determined by repeatedly deleting the node with the minimum number of neighbors (without adding new edges) until the graph is empty. The maximum number of neighbors a node had when it was deleted is the width. If the width is bounded by $i$, then variable elimination can finish the problem without introducing a new function on more than $i$ variables, *if* it does not add any new edges.

Approximate decomposition works like variable elimination, except that after eliminating a variable, it will delete newly added edges as necessary to ensure that the width of the graph remains bounded by $i$. We do not allow the algorithm to eliminate a variable with more than $i$ neighbors. However if the width limit is maintained, it will always be able to finish the problem, in the worst case by immediately deleting every new edge.

As an example, suppose approximate decomposition is run on the problem in figure 1 to compute an upper bound on the probability of the query, with an $i$ bound of 2. This means that we do not want to record a function with arity higher than 2. Starting with the moral graph in the middle, which has a width of 2, variable $F$ is eliminated first. This can be done exactly since it only has 2 neighbors. The result is shown on the right hand side of the figure. It still has width 2. The next variable to be eliminated is $A$.

---

**The Main AD Algorithm**

**Input:** Functions $F$, query variable $X_q$, operators $\otimes$ and $\odot$, complexity bound $i$.
**Output:** Bound on the value of the query.

Let $G = (V, E)$ be $F$'s graph. While $|V| > 1$:

1. Choose $X_j \in V, X_j \neq X_q$ with the smallest number of pairs of unconnected neighbors.

2. Let $F_j = \{f \in F | f \text{ mentions } X_j\}$. Set $\lambda = \otimes_{X_j} \odot_{\{f \in F_j\}} f$, $F = \lambda \cup (F - F_j)$.

3. Connect all $X_j$'s neighbors in $G$, delete $X_j$. If width$(G) > i$,

   (a) Delete new edges until width$(G) \leq i$.
   (b) Let $\{C_1, ..., C_m\}$ be the maximal cliques among $X_j$'s neighbors.
   (c) Let $\lambda_i$ have scope $C_i$, and use approximate decomposition to bound $\lambda$ by $\odot_i \lambda_i$.
   (d) Set $F = \{\lambda_i\} \cup (F - \lambda)$.

Figure 3: The Main AD Algorithm

In the top of figure 2 we show the CPTs $P(A)$, $P(B|A)$, and $P(C|A)$ which mention $A$. On the bottom left of figure 2 is the new function $\lambda(B,C) = \sum_A P(A)P(B|A)P(C|A)$ which results from eliminating $A$. This new function adds an edge between $B$ and $C$. The remaining graph is then a clique on four vertices, with a width of 3. This would exceed $i$ and is therefore not acceptable. So after computing $\lambda(B,C)$, we replace it with the product of two unary functions $\lambda_1(B)$ and $\lambda_2(C)$ which is an upper bound. In the bottom middle of the figure we show example values for $\lambda_1(B)$ and $\lambda_2(C)$, and at the bottom right we show the upper bound that they represent. The bound is not tight only in the case where $B$ is false and $C$ is true. This substitution is equivalent to deleting the edge between $B$ and $C$. After that, variable elimination can process the rest of the problem normally without violating the $i$ bound.

In general, approximate decomposition computes a bound on the value of a query variable, given $\otimes$, $\odot$, and the complexity bound $i$. Assuming that the width of the belief network's moral graph is initially no greater than $i$, it eliminates variables like variable elimination so long as the new edges do not cause the width of the graph to exceed $i$. If this does happen, then the new edge with the maximum sum of endpoint degrees is deleted until the limit is met. If $C_1, ..., C_m$ are the maximal cliques of the subgraph induced by the eliminated variable's neighbors, we want to approximate the exact elimination function $\lambda$ with $\odot_i \lambda_i$, where $\lambda_i$ has scope $C_i$. To do this, a linear program is set up and solved. The details of this step are given in the next subsection. This defines values for the bounding functions $\{\lambda_i\}$, which then replace $\lambda$, allowing variable elimination to continue. The final result is an upper or lower bound on the original query. If a bound from the other direction is desired, the algorithm must be run again from the beginning. Pseudocode for this main algorithm is given in figure 3.

## 3.2 THE APPROXIMATE DECOMPOSITION STEP

In this subsection we describe the linear programming step which is used to compute values for a set of functions $\{\lambda_1, \lambda_2, ..., \lambda_m\}$ such that their product $\lambda' = \prod_i \lambda_i$ is a good bound on $\lambda$. We will assume that $L$ is the scope of $\lambda$ and $\lambda'$. The technique needed to approximate $\lambda$ with a sum of functions is much simpler and is outlined briefly at the end.

### 3.2.1 What is a Good Bound?

We will use the notation $E_F(H(F))$ to denote $\sum_f P(F = f)H(f)$, that is, the expected value of $H$ as a function of the random variable $F$.

**Definition 1** Let $\lambda$ be a function with scope $L$, and let $F(x_L)$ be a collection of random variables, one for each assignment $x_L$ to $L$. Let $\lambda'$ be a bound on $\lambda$, also with scope $L$, and suppose $\otimes \in \{+, \max\}$. Then we define the cost of $\lambda'$ to be $C(\lambda') = E_F(|\otimes_{x_L} F(x_L) \cdot \lambda'(x_L) - \otimes_{x_L} F(x_L) \cdot \lambda(x_L)|)$.

The cost of $\lambda'$ is a measure of the error in the bound on the final query that results from substituting $\lambda'$ for $\lambda$, assuming that all previous and subsequent computation is exact. In the case of belief inference, for example, if $\lambda'$ is an upper bound on the function $\lambda$ that results from eliminating $X_k$, the query error is $\sum_{\{X_1,...,X_{k-1}\}} \prod_i f_i \lambda' - \sum_{\{X_1,...,X_{k-1}\}} \prod_i f_i \lambda = \sum_{x_L} F(x_L)\lambda'(x_L) - \sum_{x_L} F(x_L)\lambda(x_L)$, where $F(x_L) = \sum_{\{X_1,...,X_{k-1}\}-L} \prod_i f_i$ is the set of random variables summarizing our uncertainty about the rest of the problem.

In general the random variables $F(x_L)$ can encapsulate any knowledge we may have about the remainder of the problem. We will make the most conservative assumption, that nothing at all about the rest of the problem is known. The random variables then are taken to be independent, identically distributed, and uniform. The cost of the bound $\lambda'$ then depends only on $\lambda$. Theorem 1 states that the cost of $\lambda'$ for belief inference ($\otimes = \sum$) under these assumptions is $k \sum_{x_L} |\lambda' - \lambda|$, and theorem 2 states that the cost for MPE ($\otimes = \max$) is no greater than $2k \sum_{x_L} |\lambda' - \lambda|$, where $k = E_F(F(x_L))$. In the subsequent sections we will assume that the best bound for both problems minimizes the sum of absolute errors ($L_1$ distance).

**Theorem 1** Let $\lambda$ be a function on $L$, and let $\lambda'$ bound it. Let $F(x_L)$ be a collection of i. i. d. uniform random variables, and suppose $\otimes = \sum$. Then the cost $C(\lambda') = k \sum_{x_L} |\lambda'(x_L) - \lambda(x_L)|$, where $k = E_F(F(x_L))$.

**Proof:** Assume $\lambda'$ is an upper bound. When it is a lower bound the proof in analogous. Let $\epsilon = \lambda' - \lambda$. The cost is $E_F(\sum_{x_L} F(x_L) \cdot (\lambda(x_L) + \epsilon(x_L))) - E_F(\sum_{x_L} F(x_L) \cdot \lambda(x_L))$. The first expectation $E_F(\sum_{x_L} F(x_L) \cdot (\lambda(x_L) + \epsilon(x_L))) = E_F(\sum_{x_L} F(x_L) \cdot \lambda(x_L) + \sum_{x_L} F(x_L) \cdot \epsilon(x_L)) = E_F(\sum_{x_L} F(x_L) \cdot \lambda(x_L)) + \sum_{x_L} E(F(x_L)) \cdot \epsilon(x_L)$. Therefore the cost reduces to $\sum_{x_L} E_F(F(x_L)) \cdot \epsilon(x_L)$, as desired. $\square$

**Lemma 1** Let $\lambda$ be a function on $L$, and let $F(x_L)$ be a collection of i. i. d. uniform random variables each with $N$ possible values and maximum value $M$. Then the probability that $x_L$ maximizes $F(x_L)\lambda(x_L)$ is $\frac{1}{N}\sum_{\{f \in dom(F(x_L))\}} \prod_{\{y_L \neq x_L\}} \frac{f\lambda(x_L)}{M\lambda(y_L)}$.

**Proof:** The probability that $x_L$ maximizes $F\lambda$ is $\sum_{\{f \in dom(F(x_L))\}} P(F(x_L) = f) \cdot \prod_{\{y_L \neq x_L\}} P(F(y_L) \cdot \lambda(y_L) \leq f \cdot \lambda(x_L)) = \sum_{\{f \in dom(F(x_L))\}} \frac{1}{N} \cdot \prod_{\{y_L \neq x_L\}} P(F(y_L) \leq \frac{f\lambda(x_L)}{\lambda(y_L)})$. Now, assuming $F(y_L)$'s $N$ values are evenly spaced on the line between 0 and $M$, the probability that it will be less than or equal to $q$ is approximately $q/M$. So the expression for the desired probability becomes $\sum_{\{f \in dom(F(x_L))\}} \frac{1}{N} \prod_{\{y_L \neq x_L\}} \frac{f\lambda(x_L)}{M\lambda(y_L)}$, as required. $\square$

**Theorem 2** Let $\lambda$ be a function on $L$ and let $\lambda'$ be a bound on it. Let $F(x_L)$ be a set of i. i. d. uniform random variables with $N$ values and maximum value $M$, and suppose $\otimes = \max$. Then the cost $C(\lambda') \leq 2k \sum_{x_L} |\lambda'(x_L) - \lambda(x_L)|$, where $k = E_F(F(x_L))$.

**Proof:** We assume that $\lambda'$ is an upper bound (when it is a lower bound the proof is analogous). Define $\epsilon = \lambda' - \lambda$. Let $Q_L$ be the random variable ranging over assignments to $L$ that maximizes $F(x_L)\lambda'(x_L)$ and likewise let $R_L$ maximize $F(x_L)\lambda(x_L)$. Then the cost $C(\lambda') = E_F(\max_{x_L} F(x_L) \cdot \lambda'(x_L) - \max_{x_L} F(x_L) \cdot \lambda(x_L)) = E_{\{F,Q_L,R_L\}}(F(Q_L) \cdot \lambda'(Q_L) - F(R_L) \cdot \lambda(R_L)) = E_F(\sum_{x_L} P(Q_L = x_L) \cdot F(x_L) \cdot \lambda'(x_L) - \sum_{x_L} P(R_L = x_L) \cdot F(x_L) \cdot \lambda(x_L)) = E_F(\sum_{x_L} F(x_L) \cdot (P(Q_L = x_L) \cdot \lambda'(x_L) - P(R_L = x_L) \cdot \lambda(x_L))) = E_F(\sum_{x_L} F(x_L) \cdot ((P(Q_L = x_L) - P(R_L = x_L)) \cdot \lambda(x_L) + P(Q_L = x_L) \cdot \epsilon(x_L)))$.

Now, let $r(x_L) = \lambda'(x_L)/\lambda(x_L)$ be the relative error in the bound. By lemma 1, $P(R_L = x_L) = \frac{1}{N}\sum_{\{f \in dom(F(x_L))\}} \prod_{\{y_L \neq x_L\}} \frac{f\lambda(x_L)}{M\lambda(y_L)}$ and also $P(Q_L = x_L) = \frac{1}{N}\sum_{\{f \in dom(F(x_L))\}} \prod_{\{y_L \neq x_L\}} \frac{f\lambda(x_L)r(x_L)}{M\lambda(y_L)r(y_L)} \leq r(x_L) \cdot$

$\frac{1}{N} \sum_{\{f \in \text{dom}(F(x_L))\}} \prod_{\{y_L \neq x_L\}} \frac{f\lambda(x_L)}{M\lambda(y_L)}$, where the inequality follows by lower bounding $r(y_L)$ with its minimum value 1. But then $P(Q_L = x_L) \leq r(x_L) \cdot P(R_L = x_L)$, and $E_F(\sum_{x_L} F(x_L) \cdot ((P(Q_L = x_L) - P(R_L = x_L)) \cdot \lambda(x_L) + P(Q_L = x_L) \cdot \epsilon(x_L))) \leq E_F(\sum_{x_L} F(x_L) \cdot (P(R_L = x_L) \cdot (r(x_L) - 1) \cdot \lambda(x_L) + P(Q_L = x_L) \cdot \epsilon(x_L)))$. Since $\lambda r = \lambda + \epsilon$, $\epsilon = \lambda(r - 1)$, and the upper bound becomes $E_F(\sum_{x_L} F(x_L) \cdot (P(R_L = x_L) \cdot \epsilon(x_L) + P(Q_L = x_L) \cdot \epsilon(x_L))) \leq E_F(\sum_{x_L} F(x_L) \cdot 2 \cdot \epsilon(x_L)) = \sum_{x_L} 2 \cdot E_F(F(x_L)) \cdot \epsilon(x_L)$, as desired. □

### 3.2.2 Calculating a Good Bound

Assume that we are given a function $\lambda$ defined on $L$ and a set of subsets $L_1, ..., L_m$ such that $\cup_{i=1}^{m} L_i = L$. We must compute functions $\lambda_1, ..., \lambda_m$ such that $\lambda_i$ has scope $L_i$ and $\prod_i \lambda_i$ is a bound on $\lambda$ which minimizes the sum of the absolute errors. Without losing generality we will assume for the moment that it is to be an upper bound.

A straightforward way to do this is to set up a non-linear program. We introduce a variable $\lambda_i(x_{L_i})$ for every assignment $x_{L_i}$ to $L_i$, and also a non-negative variable $\epsilon(x_L)$ for every $x_L$. Now, for every $x_L$, we define a non-linear constraint $\prod_{i=1}^{m} \lambda_i(x_{L_i}) - \epsilon(x_L) = \lambda(x_L)$, where $x_{L_i}$ is consistent with $x_L$ for all $i$. Subject to these constraints, we want to minimize the objective function $\sum_{x_L} \epsilon(x_L)$, which corresponds to the $L_1$ error. The values that the optimum solution assigns to $\lambda_i(x_{L_i})$ will then be our desired result.

The problem with this approach of course is that non-linear programs are in general very difficult to solve. Instead, we will relax the constraints until they become linear, and then use the solution of the linear program as an approximation of the optimum solution. First, we introduce new variables $r(x_L)$ representing the *relative* error. We will not represent the absolute error directly, so the constraints defined above cannot be used. Instead a constraint $(\prod_{i=1}^{m} \lambda_i(x_{L_i}))/\lambda(x_L) = r(x_L)$ is generated for every $x_L$. Since we want an upper bound, $r$ must be at least 1. After taking the logarithm of both sides, this becomes $\sum_{i=1}^{m} \log(\lambda_i(x_{L_i})) - \log(\lambda(x_L)) = \log r(x_L)$. We can introduce new variables $\log \lambda_i(x_{L_i})$ and $\log r(x_L)$, and the constraints are now linear in these variables ($\log \lambda(x_L)$ of course is a constant). The constraints $r \geq 1$ become $\log r \geq 0$.

Now, for upper bounds, $r(x_L)\lambda(x_L) = \lambda(x_L) + \epsilon(x_L) = \prod_i \lambda_i(x_{L_i})$, so $\epsilon(x_L) = \lambda(x_L)(r(x_L) - 1)$. We want to minimize $\sum_{x_L} \epsilon(x_L)$, but unfortunately the only relevant variables that have meaning under the constraints are $l_r(x_L) = \log r(x_L)$. The objective function is therefore $\sum_{x_L} \lambda(x_L)(\exp(l_r(x_L)) - 1)$. This is

```
Minimize
 .23 l_r(B̄, C̄) + .15 l_r(B̄, C) + .33 l_r(B, C̄) + .29 l_r(B, C)
subject to:
 log λ_1(B̄) + log λ_2(C̄) - l_r(B̄, C̄) = log λ(B̄, C̄) = -0.635
 log λ_1(B̄) + log λ_2(C) - l_r(B̄, C) = log λ(B̄, C) = -0.83
 log λ_1(B) + log λ_2(C̄) - l_r(B, C̄) = log λ(B, C̄) = -0.484
 log λ_1(B) + log λ_2(C) - l_r(B, C) = log λ(B, C) = -0.535
 l_r(B̄, C̄), l_r(B̄, C), l_r(B, C̄), l_r(B, C) ≥ 0
```

Figure 4: The linear program computing the bounding functions in figure 2

```
Procedure Approximate Decomposition
Input: Function λ with scope L, subsets
L_1, ..., L_m such that ∪_i L_i = L.
Output: Set of functions λ_1, ..., λ_m such that λ_i
has scope L_i and ∏_i λ_i is an upper bound on λ
(lower bound is similar).

 1. Begin constructing a new linear program:
    (a) Introduce linear program variables
        log λ_i(x_{L_i}) and l_r(x_L) ≥ 0.
    (b) ∀λ(x_L) ≠ 0, introduce the constraint
        ∑_i log λ_i(x_{L_i}) - log λ(x_L) = l_r(x_L).
    (c) ∀λ(x_L) = 0, introduce the constraint
        ∑_i log λ_i(x_{L_i}) - Z ≤ l_r(x_L).
    (d) Let the objective be to minimize
        ∑_{x_L} max(10^-5, λ(x_L)/∑_{x_L} λ(x_L)) l_r(x_L).

 2. Solve the program using a standard LP
    algorithm.  Define {λ_i} with the solution.
```

Figure 5: The Approximate Decomposition Procedure

nonlinear since it involves the exponential function. Therefore we will approximate $\exp(l_r) - 1$ as $dl_r + c$, for appropriate constants $d \geq 0$ and $c$. The objective function then becomes $\sum_{x_L} \lambda(x_L)(dl_r(x_L) + c) = c' + d \sum_{x_L} \lambda(x_L) l_r(x_L)$ for some constant $c'$. The solution that minimizes this also minimizes $\sum_{x_L} \lambda(x_L) l_r(x_L)$, so we will use the latter as our approximate objective function. The linear program is then complete.

If we want a lower bound on $\lambda$ instead, then the constraints are $r(x_L) = \lambda(x_L)/(\prod_i \lambda_i(x_{L_i}))$, so $\sum_i \log \lambda_i - \log \lambda = -\log r$. Now, $\prod_i \lambda_i(x_{L_i}) = \lambda(x_L)/r(x_L) = \lambda(x_L) - \epsilon(x_L)$, so $\epsilon(x_L) = \lambda(x_L)(1 - 1/r(x_L)) = \lambda(x_L)(1 - 1/\exp(l_r(x_L)))$. As before, we approximate $(1 - 1/\exp(l_r(x_L)))$ with $dl_r(x_L) + c$ for appropriate constants $d \geq 0$ and $c$, and the objective function again becomes $\sum_{x_L} \lambda(x_L) l_r(x_L)$.

As an example, consider figure 2 in subsection 3.1. The linear program that was used to compute the bounding functions $\lambda_1$ and $\lambda_2$ is given in figure 4.

There is one issue we have not addressed so far, namely how to deal with cases when $\lambda(x_L)$ is 0. Then $\log \lambda(x_L)$ is negative infinity, which does not allow us to set up the program as described above. Instead, we approximate $\log 0$ as a large negative number $Z$ (say -40). Then, for an upper bound we introduce the constraint $\sum_i \log \lambda_i - Z \le \log r$, and for a lower bound we have $\sum_i \log \lambda_i \le Z$. In the second case, $\prod_i \lambda_i$ will not exceed $10^Z$, but all the factors in the product will be nonzero (although there will generally be at least one very small one). To enforce the lower bound in this case, after the linear program has been solved, we set the $\lambda_i$ which was assigned the smallest value directly to 0.

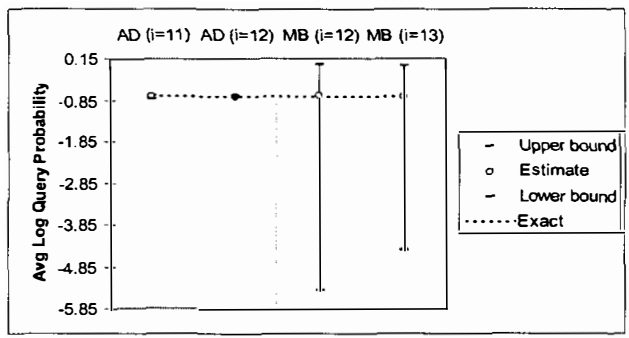

Figure 6: CPCS2 Belief Inference (360 binary variables, $w^* = 20$, 5 observations)

As the log of the relative error of the bounding functions on a given domain value goes to infinity, the ideal cost function for upper bounds increases exponentially, and the ideal cost function for lower bounds approaches a constant. The approximation in both cases is linear, meaning that for upper bounds it will be an arbitrarily bad underestimation and for lower bounds it will be an overestimation. Therefore the linear objective function is only meaningful when the relative errors are more or less bounded. If one cost coefficient in this function is very small in relation to the others, the optimum LP solution might assign the corresponding relative error a very high value. To avoid this, in our implementation we normalized all of the cost coefficients so that they added up to one, and then set any coefficient less than $10^{-5}$ to $10^{-5}$. This substantially increased the accuracy in our experiments.

Finally it is interesting to note that the objective function for lower bounds $\sum_{x_L} \lambda(x_L) l_r(x_L) = \sum_{x_L} \lambda(x_L)(\log(\lambda(x_L)/\prod_i \lambda_i(x_{L_i})))$ is actually the KL distance between the target function and its bound, assuming both are normalized probabilities. A similar observation holds for upper bounds. In general however the intermediate functions produced by variable elimination and their bounds are not normalized.

Pseudocode for the approximate decomposition procedure is given in figure 5.

|        | AD (i=11)  |          | MB (i=12)  |          |
|--------|-----------|----------|-----------|----------|
|        | Qry. Pr.  | Evd. Pr. | Qry. Pr.  | Evd. Pr. |
| Low    | -0.825    | -3.65    | -5.41     | -7.12    |
| Est.   | -0.771    | -3.63    | -0.761    | -3.56    |
| High   | -0.74     | -3.6     | -0.00551  | -2.48    |
| Exact  | -0.775    | -3.64    | -0.775    | -3.64    |
| Est. $\epsilon$ | 0.00493 | 0.0159 | 0.015 | 0.284 |
| Hi-Lo  | 0.0854    | 0.046    | 5.41      | 4.63     |
| Time   | 10.8s     |          | 7.75s     |          |
|        | AD (i=12) |          | MB (i=13) |          |
|        | Qry. Pr.  | Evd. Pr. | Qry. Pr.  | Evd. Pr. |
| Low    | -0.783    | -3.64    | -4.45     | -6.27    |
| Est.   | -0.765    | -3.64    | -0.772    | -3.49    |
| High   | -0.749    | -3.64    | -0.0303   | -2.59    |
| Exact  | -0.775    | -3.64    | -0.775    | -3.64    |
| Est. $\epsilon$ | 0.0111 | 0.00103 | 0.00433 | 0.241 |
| Hi-Lo  | 0.034     | 0.00292  | 4.42      | 3.68     |
| Time   | 15.6s     |          | 13.6s     |          |

Table 1: CPCS2 Belief Inference (360 binary variables, $w^* = 20$, 5 observations, exact algorithm takes 202s)

|        | AD ($i=11$) | MB ($i=12$) | AD ($i=12$) | MB ($i=13$) |
|--------|------------|------------|------------|------------|
| Low    | -12.6      | -24.1      | -12.5      | -22.3      |
| Est.   | -12.3      | -15.6      | -12.3      | -15.4      |
| High   | -12.1      | -11.8      | -12.2      | -12        |
| Exact  | -12.3      | -12.3      | -12.3      | -12.3      |
| Est. $\epsilon$ | 0.223 | 3.36 | 0.125 | 3.17 |
| Hi-Lo  | 0.527      | 12.3       | 0.314      | 10.3       |
| Time   | 15.4s      | 8.43s      | 20.5s      | 14.5s      |

Table 2: CPCS2 MPE (360 binary variables, $w^* = 20$, 5 observations, exact algorithm takes 205s)

### 3.2.3 Computing a Bound for MAX-CSP

If $\lambda$ is to be bounded by the sum of the $\lambda_i$'s, instead of the product, the task becomes much easier. For an upper bound, we introduce a constraint $\sum_i \lambda_i(x_L) - \epsilon(x_L) = \lambda(x_L)$ for all $x_L$. The lower bound is similar. Notice that there is no need to take the logarithms, and cases where $\lambda(x_L)$ is zero do not need to be handled specially. The objective function to be minimized is then the exact sum of absolute errors, $\sum_{x_L} \epsilon(x_L)$.

## 4 EMPIRICAL RESULTS

We compared the approximations computed by approximate decomposition with those of mini buckets on a number of problems. For each experiment, random evidence and a random query variable was selected, and bounds and an estimate on the value of the query were computed by each algorithm. The results reported in all cases are the average of 25 experiments.

To solve the linear programming problems, we used ILOG CPLEX's primal simplex optimizer. AD effectively solved each problem twice, once for the upper bound $U$, and once for the lower bound $L$. For proba-bilistic queries, the estimate was $\exp(\frac{\log U + \log L}{2})$. For MAX-CSP, the estimate was $\frac{U+L}{2}$. This was effective since the bounds were generally tight and well centered around the exact answer.

MB processed each problem 3 times, twice for the bounds, and once for the estimate. A lower bound on the query was found by projecting the bucket variable out of all of the mini buckets beyond the first in the bucket with the min operator. Likewise max was used for the upper bound and mean for the estimate. MB's bounds were loose and generally poorly centered around the exact, so AD's technique was not used. For details about mini buckets see [Dechter and Rish, 1997].

In table 1, we report the results of 25 belief inference experiments on the CPCS2 network, each with five observations. AD took longer than MB with the same $i$ bound because of the overhead of linear programming, so MB was given an increased $i$ bound to make the times comparable. "Qry. Pr." describes the approximation of the conditional probability of the query, and "Evd. Pr." describes the approximation of the probability of the evidence. The conditional probability can be bounded in a straightforward way from bounds on the joint, which was what the algorithms computed directly. Each entry in the table represents the average base 10 logarithm of the noted value, averaged over all 25 experiments, and in the case of the conditional probability, over the query domain values also. The rows "Low", "Est.", and "High" give the lower bounds, estimates, and upper bounds. "Exact" is the exact answer, and "Est. $\epsilon$" is the factor by which the estimate diverged from the exact. "Hi-Lo" is the upper bound divided by the lower bound, and "Time" is the number of seconds required by each algorithm (not the logarithm in this case).

For all the experiments, the main measures of approximation accuracy are Est. $\epsilon$ and Hi-Lo. In table 1, under the query probability column, Est. $\epsilon$ is 0.00493 for AD at $i = 11$. This means that the estimate of the conditional probability diverged from the exact by a factor of $10^{0.00495} = 1.01$ in the average case. Hi-Lo is 0.0854. That means that the average upper bound was higher than the lower bound by a factor of $10^{0.0854} = 1.21$. High is $-0.74$, meaning that the average upper bound was $10^{-0.74} = 0.18$. By contrast, for MB at $i = 13$ approximating the same quantity, the average error in the estimate was $10^{0.00433}$, just slightly better than AD's estimates, but the average upper bound was a factor of $10^{4.42} = 26,302$ higher than the lower. Clearly MB was not effective at computing bounds in this case. In figure 6, for comparison, we have graphically displayed the Low, Est., High, and Exact values for the query probabilities. Each column corresponds to an algorithm and an $i$ bound. In a column, the top bar is High, the bottom is Low, the circle is Est., and the dotted line is Exact.

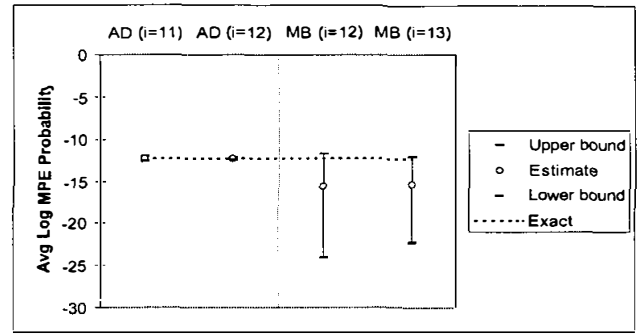

Figure 7: CPCS2 MPE (360 binary variables, $w^* = 20$, 5 observations)

|  | AD (i=12) | | MB (i=14) | |
| --- | --- | --- | --- | --- |
|  | Qry. Pr. | Evd. Pr. | Qry. Pr. | Evd. Pr. |
| Low | -1.77 | -4.52 | -9.42 | -10.5 |
| Est. | -1.12 | -4.21 | -1.09 | -4.03 |
| High | -0.772 | -3.89 | 0 | -2.17 |
| Exact | -1.1 | -4.24 | -1.1 | -4.24 |
| Est. $\epsilon$ | 0.0402 | 0.122 | 0.0238 | 0.255 |
| Hi-Lo | 1 | 0.639 | 9.42 | 8.3 |
| Time | 30.9s | | 28.9s | |

Table 3: CPCS3 Belief Inference (422 binary variables, $w^* = 22$, 5 observations, exact algorithm takes 1020s)

In table 2 we report the results of MPE experiments on CPCS2. For each value of the query variable, we approximated the probability of the most probable explanation consistent with it and the evidence, then averaged the logs of the approximations over all domain values. AD at $i = 11$ computed an upper bound that was a factor of $10^{0.527} = 3.37$ higher than the lower in the average case, whereas MB's upper bound at $i = 13$ was $10^{10.3}$ times higher. Unlike in the belief inference case, its estimate was also significantly inaccurate. AD's estimate diverged from the exact by an average factor of $10^{0.223} = 1.67$, while MB's error factor was $10^{3.17} = 1,479$. As before, in figure 7 we graphically display the values of the table. Note that MB's intervals are very poorly centered around the exact and that the estimate is far off.

In table 3 we report the results of belief inference on the CPCS3 network. Again AD computed very sharp bounds on the conditional and evidence probabilities, whereas MB's were too loose to be useful. The estimates produced by both algorithms were both quite accurate. In table 4, we report the results of the MPE task on CPCS3. AD's bounds and estimate were orders of magnitude better than MB's in all cases.

In table 5 we report the results of belief inference on a class of large random networks. Each network's graph had 115 variables, where 110 nodes had 3 random parents and the other five were roots. These networks were very difficult, with an average induced width $w^*$ of 34, well beyond the reach of an exact algorithm. AD's bounds were much tighter than MB's. Of course the error in the estimates could not be checked. But they agreed with each other very closely, as in the other belief inference tasks.

|  | AD (i=11) | MB (i=13) | AD (i=12) | MB (i=14) |
|---|---|---|---|---|
| Low | -13.6 | -30.6 | -14.3 | -27.2 |
| Est. | -13 | -17 | -13.2 | -16.1 |
| High | -12.5 | -12 | -12.1 | -12 |
| Exact | -12.8 | -12.8 | -12.8 | -12.8 |
| Est. $\epsilon$ | 0.39 | 4.22 | 0.618 | 3.26 |
| Hi-Lo | 1.06 | 18.6 | 2.18 | 15.2 |
| Time | 25.6s | 17.9s | 39.9s | 30.9s |

Table 4: CPCS3 MPE (422 binary variables, $w^* = 22$, 5 observations, exact algorithm takes 1020s)

|  | AD (i=11) | | MB (i=14) | |
|---|---|---|---|---|
|  | Qry. Pr. | Evd. Pr. | Qry. Pr. | Evd. Pr. |
| Low | -2.75 | -2.87 | -15.4 | -12.2 |
| Est. | -0.314 | -1.65 | -0.31 | -1.57 |
| High | 0 | -0.436 | 0 | 0 |
| Hi-Lo | 2.75 | 2.43 | 15.4 | 12.2 |
| Time | 27s | | 21.5s | |

Table 5: Random Belief Network Inference (115 binary variables, average $w^* = 34$, 5 observations, exact algorithm intractable)

Finally in table 6, we report some results on a class of random MAX-CSP problems. Each instance had 30 ternary variables and 125 random binary constraints, each of which disallowed half of all possible value pairs. The entries in the table are not logarithms, but direct approximations of the number of constraints violated by the optimum solution. In this case also AD substantially outperformed MB both in terms of the estimation quality and the tightness of the bounds.

To conclude, in our experiments AD was much more effective than MB at computing bounds on probabilistic and deterministic queries, and also produced substantially more accurate estimates in the case of the MPE task and MAX-CSP. For belief inference, the estimates of both algorithms were very accurate when they could be compared to the exact answer.

## 5 CONCLUSION

In this paper, we introduced a method for bounding and estimating probabilistic and deterministic queries, called approximate decomposition. It works by bounding a large complex function with a collection of smaller and simpler ones. We showed what properties the bound should have in the ideal case and how that ideal could be tractably approximated with linear programming. In the future, we plan to investigate the idea of a multiple-query version of the algorithm, along the lines of clique tree propagation.

### Acknowledgements

Thanks to James D. Park for pointing out the connection between the linear objective function and the KL distance. This work was supported in part by the NSF grant IIS-0086529 and MURI ONR award N00014-00-1-0617.

|  | AD (i=7) | MB (i=8) | AD (i=8) | MB (i=9) |
|---|---|---|---|---|
| Low | 17 | 14.6 | 17.8 | 15.7 |
| Est. | 20.6 | 25.8 | 20.6 | 24.5 |
| High | 24.2 | 35.7 | 23.4 | 32.5 |
| Exact | 21.3 | 21.3 | 21.3 | 21.3 |
| Est. $\epsilon$ | 1.02 | 4.45 | 1.01 | 3.15 |
| Hi-Lo | 7.21 | 21.1 | 5.58 | 16.8 |
| Time | 7.01s | 9.27s | 34.3s | 22.8s |

Table 6: MAX-CSP (30 ternary variables, 125 binary constraints, average $w^* = 14$, exact algorithm takes 1190s)